\documentclass[letterpaper, 10 pt, conference]{ieeeconf}
\IEEEoverridecommandlockouts
\usepackage{cite}
\usepackage{amsmath,amssymb,amsfonts}
\usepackage{algorithmic}
\usepackage{graphicx}
\usepackage{textcomp}
\usepackage{xcolor}
\usepackage{makecell}
\usepackage{bm}
\usepackage{subfigure}
\usepackage{wasysym}

\usepackage{pifont}
\newcommand{\cmark}{\ding{51}}%
\newcommand{\xmark}{\ding{55}}%

\newcommand{\qd}{\dot{\bm{q}}}
\newcommand{\inR}[1]{\in \mathbb{R}^{#1}}

\def\BibTeX{{\rm B\kern-.05em{\sc i\kern-.025em b}\kern-.08em
    T\kern-.1667em\lower.7ex\hbox{E}\kern-.125emX}}

\usepackage{flushend}

\begin{document}

\title{\LARGE \bf
Simultaneous Pick and Place Detection by Combining\\SE(3) Diffusion Models with Differential Kinematics
}

\author{Tianyi Ko$^{*\dagger}$, Takuya Ikeda$^*$, Bal\'azs Opra$^*$, Koichi Nishiwaki$^*$
\thanks{$*$ T. Ko, T. Ikeda, B. Opra, and K. Nishiwaki are with Woven by Toyota, Inc., 3-2-1 Nihonbashi-Muromachi, Chuo-ku, Tokyo, Japan.}
\thanks{$\dagger$ Corresponding author. \texttt{tianyi.ko@woven.toyota}}
}

\maketitle
\thispagestyle{empty}
\pagestyle{empty}

\begin{abstract}
Grasp detection methods typically target the detection of a set of free-floating hand poses that can grasp the object.
However, not all of the detected grasp poses are executable due to physical constraints.
Even though it is straightforward to filter invalid grasp poses in the post-process, such a two-staged approach is computationally inefficient, especially when the constraint is hard.
In this work, we propose an approach to take the following two constraints into account during the grasp detection stage, namely,
(i) the picked object must be able to be placed with a predefined configuration without in-hand manipulation (ii) it must be reachable by the robot under the joint limit and collision-avoidance constraints for both pick and place cases.
Our key idea is to train an SE(3) grasp diffusion network to estimate the noise in the form of spatial velocity, and constrain the denoising process by a multi-target differential inverse kinematics with an inequality constraint, so that the states are guaranteed to be reachable and placement can be performed without collision.
In addition to an improved success ratio, we experimentally confirmed that our approach is more efficient and consistent in computation time compared to a naive two-stage approach.

\end{abstract}

\section{Introduction}
\label{sec:introduction}
Pick-and-place is one of the most fundamental applications of robots.
Despite the significant number of works on generating ``pick" poses, limited works focus on simultaneously considering both picking and placing.
A single robot arm with a simple hand often leaves no margin for in-hand manipulation or handover capability.
In such cases, a large portion of grasps that are valid for picking are not valid for placing, taking into account the arm's reachability and collision with the environment.
For example, if the task is uprighting a bottle, the robot should not grasp the bottle's bottom because even though the grasp is valid, it would not enable collision-free placement.
Furthermore, in many cases the target placement pose has multiple choices, e.g., ``standing upright" allows arbitrary rotation around the bottle's longitudinal axis.
A naive workflow is to first sample a grasp only considering the hand, then check the arm's reachability; if it is reachable, perform a grid search for the placement.
When either pick or place conditions are challenging in terms of reachability or collision avoidance, such a process is highly inefficient.
In this work, we propose a reverse workflow: start from reachable and collision-free configurations and update the pick-place configurations based on differential kinematics guided by a SE(3) diffusion model that estimates the hand's spatial velocity.
In addition, we treat the placement pose as a decision variable to support diverse applications.

Our problem setting assumes multiple objects, not included in the training data, piled on the tabletop in a cluttered manner.
A single depth image of the scene is provided accompanied by the target object's segmentation mask and 6D pose.
The target placement pose is provided as a cost function, e.g., a single desired SE(3) pose, a set of affordable discrete SE(3) poses, or a continuous set, such as allowing for arbitrary rotation around the gravity direction.
Our target is to generate a dense trajectory in the configuration space, including the motion of approaching the object, lifting, moving to the location of the place, and retracting the hand.

Our contribution is twofold.
On the grasp detection side, we propose an SE(3) grasp detection diffusion model that is jointly trained with a feedforward type grasp detector and uses the feedforward network output to validate the diffusion model output.
On the pick-place pipeline side, we propose to solve a multi-target differential inverse kinematics (Diff-IK) with the grasp diffusion model's output as one of the tasks.

\begin{figure}[tbp]
    \centerline{\includegraphics[width=1.0\columnwidth]{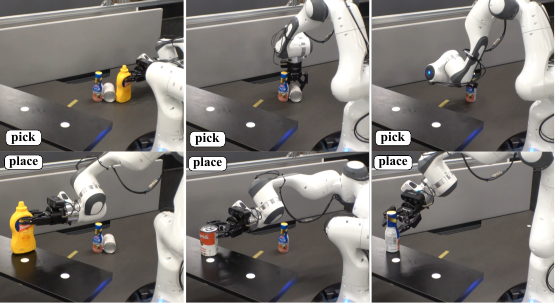}}
    \caption{
        Assuming the 6D pose and segmentation mask of the target object is provided,
        our approach simultaneously optimizes both pick and place pose in the dense joint trajectory form.
        }
    \label{fig:experiment}
\end{figure}

\section{Related Works}
The recent decade has seen the emergence of data-driven grasp synthesis~\cite{grasp_detection_survey, dexnet2, gpd_ijrr, s4g, regnet, gsnet, contact_graspnet, edge_grasp, vgn, giga, pvgn, pvgn_bingham}, typically referred to as \textit{grasp detection}.
Pioneering works include 2D grasp detection by Mahler \textit{et al.}~\cite{dexnet2} and 3D grasp detection by Ten Pas \textit{et al.}~\cite{gpd_ijrr}.
While a large number of works~\cite{s4g, regnet, gsnet, contact_graspnet, edge_grasp} take a point cloud as input, some works~\cite{vgn, giga, pvgn} opt for voxel representation.
Breyer \textit{et al.}~\cite{vgn} represented the scene as a truncated signed distance function (TSDF) and employed a 3D fully convolutional network to estimate per-voxel grasp probability.
Our work is related to this because we use the same scene encoder to condition our diffusion model and show that jointly training a VGN~\cite{vgn} decoder improves the diffusion model.

Whereas the majority of works estimate grasp probability, Ko \textit{et al.}~\cite{pvgn} proposed to estimate the gravity-rejection score, which is the magnitude of gravity direction disturbance a grasp can support, to prioritize physically robust grasps and realized detecting power grasps~\cite{cutkosky1989grasp}.
Our work also adopts this idea, but as ours is a generative model, we use this score to condition the diffusion model.
The model first tries to generate only highly robust grasps, and only when it fails does it try to generate modestly robust grasps.

\begin{figure*}[tbp]
    \centerline{\includegraphics[width=0.9\linewidth]{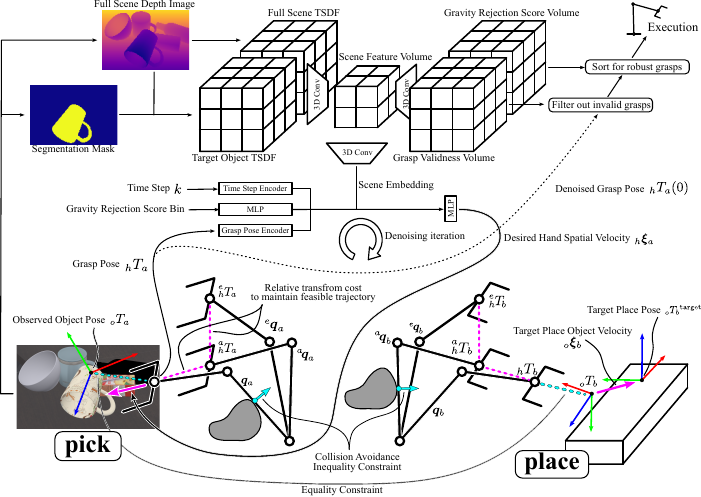}}
    \caption{
    Inference pipeline of the proposed approach.
    The upper half illustrates the network architecture.
    It is conditioned by scene observation, one from the whole scene and another from the segmented scene.
    At each timestep, it takes the hand's SE(3) pose as the input and outputs the desired hand spatial velocity.
    The bottom half illustrates the state update constrained by the Diff-IK guided by the network output.
    }
    \label{fig:diagram}
\end{figure*}

To jointly solve grasp detection with other tasks, the grasp detector should be differentiable with regard to the hand's SE(3) pose.
Wang \textit{et al.}~\cite{wang2019manipulation} first acquired initial grasp poses based on geometry-based grasp synthesis~\cite{graspit} and applied iterative surface fitting (ISF), which fits the 3D shapes of the hand and the object, so that the grasp can be jointly optimized with an optimization-based motion planner CHOMP~\cite{chomp}.
Weng \textit{et al.}~\cite{ngdf} extended \cite{wang2019manipulation} by replacing the model-based grasp synthesizer with the neural grasp distance function (NGDF).
By leveraging the auto-differentiation of NGDF, they could also omit the ISF.
Urain \textit{et al.}~\cite{se3_diffusionfield} further extended \cite{ngdf} by introducing an SE(3) diffusion model instead of NGDF.
Choi \textit{et al.}~\cite{choi2024towards} first detected grasp poses by a feed-forward grasp detector~\cite{contact_graspnet} and fitted a continuous Gaussian process distance field so that the grasp detection is combined with an online tracking controller to grasp a moving object.
The largest difference between our work and that of those prior works is that we combine picking and placing into a single framework, while those works only focus on picking.
Furthermore, while \cite{wang2019manipulation} requires explicit object mesh and \cite{ngdf, se3_diffusionfield} require the known-category object to be physically isolated, our approach can handle unseen objects placed in a cluttered manner.
While \cite{choi2024towards} tries to smooth the output of a feed-forward grasp detector, thus the performance is bounded by the feed-forward grasp detector, our work shows that jointly training a diffusion network with a feed-forward grasp detector can outperform the original feed-forward one even in the success ratio.

Despite numerous works on generative methods for grasp detection~\cite{6dofgraspnet, se3_diffusionfield}, it is reported that generative methods generally underperform feed-forward architectures~\cite{implicit_grasp_diffusion}.
To fill this gap, Song \textit{et al.}~\cite{implicit_grasp_diffusion} and Zhang \textit{et al.}~\cite{dexgraspnet2} concurrently proposed to use local features, rather than the global feature extracted from the whole scene, for their diffusion models and reported a better score than the feed-forward type architecture even for cluttered scenes.
Our work is related to these works in that we use non-global features.
However, since we leverage object-centric features, more specifically TSDF from the segmented depth image, our model can handle SE(3) pose including the translation.
Local features at a set of~\cite{implicit_grasp_diffusion} or a single~\cite{dexgraspnet2} key point, on the other hand, can only handle SO(3) rotation in a fixed location~\cite{implicit_grasp_diffusion} or small translation correction around a fix point~\cite{dexgraspnet2}.

Generating stable and feasible placing poses is also an active research field.
Paxton \textit{et al.}~\cite{paxton2022predicting} proposed an approach to generate unseen objects' placement poses based on semantic context.
Liu \textit{et al.}~\cite{structdiffusion} proposed a diffusion model that is conditioned by language to generate unseen objects' placement poses.
Simeonov \textit{et al.}~\cite{shelving_stacking} also proposed a diffusion model that is suitable for highly multimodal arrangement candidates.
Mishra \textit{et al.}~\cite{reorientdiff} focused on generating an intermediate placement pose, which is commonly referred to as a reorientation/grasping problem.
All these works, in addition to manually specifying placement poses, can be used to generate ``target placement set" as the input to our method.

\section{Method}
\subsection{An SE(3) Grasp Diffusion Model Jointly Trained with a FeedForward Grasp Detector}
\label{sec:diffusion}
This section introduces a grasp diffusion model~\cite{ddpm} that iteratively denoises hand poses in SE(3) by estimating the noise in the form of spatial velocity $\bm{\xi} = [\bm{w}^\top \bm{v}^\top]^\top$, where  $\bm{w}, \bm{v} \inR{3}$ is the angular and translational velocity respectively.
The hand pose is updated by applying $\bm{\xi}$ for a time delta $\Delta t$, where $\Delta t$ is a predefined hyperparameter and in this work we set it as 1 s.
The upper half of Fig.~\ref{fig:diagram} illustrates the network architecture.
The model is conditioned by the scene observation as TSDF volumes~\cite{sdf}.
In order to encode both scene-level features and object-centric features, we create two TSDF volumes: one from the full depth image and another from the depth image masked by the segmentation mask of the target object.
As a result, the input for the conditioning encoder has a shape of $2\times D\times D\times D$ where $D=40$ is the voxel resolution aligned with other works~\cite{vgn, giga, pvgn, pvgn_bingham}.
This is processed by a 3D fully convolutional encoder proposed in \cite{vgn}.
To condition the diffusion model, this scene feature volume is further processed by a 3D convolution layer to acquire a 128-sized scene embedding and concatenated with the grasp pose embedding.

We encode an SE(3) grasp pose by the locations of multiple points attached to the hand frame.
The distance between the points corresponds to the sensitivity ratio for translation and rotation.
In this work, we selected nine points in a two-meter square cube with its center aligned with the hand's tool center point (TCP), resulting in $3 \times 9$ input, which is processed by a MLP for the grasp pose embedding.
We used Gaussian Fourier Projection~\cite{deepsdf} for the time embedding and the MLP backbone identical to \cite{se3_diffusionfield}, while we changed the output channel from the original one for the energy to six for the spatial velocity.

Ko \textit{et al.}~\cite{pvgn} reported that prioritizing physically robust grasps is advantageous in grasping heavy and rigid objects with power grasping~\cite{cutkosky1989grasp} by underactuated hands~\cite{handbook_hand}.
While \cite{pvgn} estimates the gravity rejection score as the output, we use the score to condition the network.
We split the grasps into three \textit{gravity rejection score bin}s, namely, grasps that can only support under 15 N gravity force, those that can support between 15 N and 30 N, and those that can support more than 30 N, and represent the bin as a one-hot vector for the network input.
At inference time, we first condition the network to generate grasps in the highest gravity rejection score bin, because they are the safest choices for unknown objects.
However, such grasps are not always possible in cluttered scenes.
For example, a high gravity rejection score typically requires the fingers to wrap around the object to form a power grasp, but it may be infeasible if another object is standing next to the target.
In such cases, medium or low gravity rejection score bins are still useful because precision grasps typically have less chance of collision.
We therefore also try to generate grasps with lower score bins once generation conditioned by a higher score bin fails.

The scene feature volume is not only used to condition the diffusion model, but also passed to a 3D fully convolutional decoder similar to \cite{vgn, pvgn} to estimate two grasp volumes, namely a grasp validness volume representing per-voxel grasp probability and a gravity rejection score volume representing per-voxel gravity rejection score.
As described in the following section, our denoising process is constrained to stay within the robot's reachable manifold.
This leads to a risk of local minima where the denoised grasp pose is not feasible for a successful grasp.
We therefore project the TCP of the denoised grasp poses to the grasp validness volume to filter out invalid grasps.
Even if conditioned by the same gravity rejection score bin, multiple denoised grasp poses may have different gravity rejection scores.
In order to prioritize the most robust grasp, we use the gravity rejection score volume to sort the denoised grasp poses in the same batch for execution.

We follow the process by \cite{pvgn} for data generation.
(i) Create initial per-object grasp pose candidates based on antipodal grasps.
Randomize the grasp poses and perform grasp simulations in gravity-less simulation to get per-object stable grasps with both precision and power grasping modes.
(ii) Apply multiple direction disturbances with increasing magnitude for each grasp in gravity-less simulation to acquire the multi-dimension disturbance rejection score.
(iii) Create random and cluttered scenes.
Project per-object grasp poses to the scene with colliding grasp poses removed.
Project the disturbance rejection scores to the scene's gravity direction to acquire the gravity rejection scores.
In total, we create 5K scenes consisting of 56 kinds of household object meshes.
The training typically takes 24 - 48 hours with a machine with 8 Nvidia V100 GPUs.

\subsection{Denoising Process Constrained by Differential IK}
This section introduces a denoising process constrained by the robot kinematics (see Fig.~\ref{fig:diagram} bottom for illustration).
For compact notation, we denote the cases of pick and place by subscription $\alpha$ and $\beta$, respectively, and $i \in \{\alpha, \beta\}$ to represent both cases.
For simplicity, we first discuss the pick-only case.
We start from a heuristic collision-free initial joint position $\bm{q}_\alpha(k=K) \inR{n}$ where $n$ denotes the number of joints of the robot, and iteratively update the joint position by $\bm{q}_\alpha(k-1) = \bm{q}_\alpha(k) + \qd_\alpha(k)\Delta t$.
Note that here the timestep $k$ is for conditioning the diffusion model (in this work we set $K=100$) and thus does not have physical meaning, while time delta $\Delta t$ has a unit of second to convert velocity to displacement (in this work, we set $\Delta t = 1$).
Both $K$ and $\Delta t$ need to be consistent with the one used in \ref{sec:diffusion}.
In the following discussions we omit $k$ for simple notation.
At each timestep, we solve the forward kinematics to acquire the hand pose $_h T_\alpha \inR{4\times4}$ ($h$ stands for ``hand").
This is fed into the diffusion model together with the timestep $t$, which outputs the target hand velocity $_h \bm{\xi}_\alpha$.
This leads to a simple quadratic programming (QP) based Diff-IK~\cite{russ_manipulatio}:
\begin{gather}
\label{eq:pick}
    \min_{\qd_\alpha} ||_h \bm{\xi}_\alpha - J_\alpha \qd_\alpha||^2~~s.t. \\
    \bm{q}_\mathrm{min} - \bm{q}_\alpha \leq \qd_\alpha \Delta t \leq  \bm{q}_\mathrm{max} - \bm{q}_\alpha \nonumber \\
    _cJ_\alpha \qd_\alpha \leq 0. \nonumber
\end{gather}
The first constraint denotes the joint limit, with $\bm{q}_\mathrm{min}, \bm{q}_\mathrm{max}$ being the lower and upper limits.
We apply the second constraint only when any of the robot links are closer to the environment than a threshold, where $_cJ \inR{m\times n}$ is the contact jacobian with $m$ denoting the number of contacts.
Note that the ``robot'' includes the hand also, and the ``environment'' includes primitive shapes extraced from the initial observation (in this work the table top) and point cloud of the objects aquired at each observation.

To consider the placement, we introduce the placement joint position $\bm{q}_\beta \inR{n}$ and iteratively update it by computing $\qd_\beta$.
At each time step, we get the placed object pose $_o T_\beta \inR{4\times4}$ ($o$ stands for ``object") so that $_hT_\beta^{-1} {_oT_\beta} = {_hT_\alpha^{-1}} {_oT_\alpha}$ stands.
Here, $_oT_\alpha$ is the pose of the target object from the observation.
This formulates that the relative transformation between the hand and the object must be consistent for pick and place because we assume that the object doesn't move in the hand.
In order to let $_oT_\beta$ converge to the target place pose $_oT_\beta{^\mathrm{target}}$, we compute the target place object velocity $_o\bm{\xi}_\beta = [_o\bm{w}_\beta^\top, {_o\bm{v}}_\beta^\top]^\top$ based on their pose difference.
If multiple choices of place pose were given, $_o\bm{\xi}_\beta$ can be selected as the minimum norm one.
By adding another decision variable $\bm{\xi}_\beta = [\bm{w}_\beta^\top, {\bm{v}}_\beta^\top]^\top$, the problem in Eq.~\ref{eq:pick} is now extended as
\begin{gather}
\label{eq:place}
    \min_{\qd_i, \bm{\xi}_\beta} ||_h \bm{\xi}_\alpha - J_\alpha \qd_\alpha||^2 + ({_o\bm{\xi}}_\beta - \bm{\xi}_\beta)^\top Q_\beta ({_o\bm{\xi}}_\beta - \bm{\xi}_\beta)~~s.t. \\
    {_oR}_\alpha^\top J_\alpha^w\qd_\alpha = {_oR}_\beta^\top (J_\beta^w\qd_\beta - \bm{w}_\beta) \nonumber \\
    {_oR}_\alpha^\top J_\alpha^v\qd_\alpha = {_oR}_\beta^\top (J_\beta^v\qd_\beta - \bm{v}_\beta - \bm{w}_\beta \times ({_h\bm{p}}_\beta - {_o\bm{p}}_\beta)) \nonumber \\
    \bm{q}_\mathrm{min} - \bm{q}_i \leq \qd_i\Delta t \leq  \bm{q}_\mathrm{max} - \bm{q}_i \nonumber \\
    _cJ_i \qd_i \leq 0. \nonumber
\end{gather}
Here, $Q_\beta \in \mathrm{diag}(6)$ is placing weight matrix.
If placing with any rotation around the z-axis is accepted, e.g., uprighting a bottle, we can set $Q_\beta[3, 3] = 0$.
This problem is typically termed an incomplete orientation constraint~\cite{ioc_tracking}.
Similarly, we can also allow arbitrary horizontal translation inside a region, e.g., anywhere on a table, by setting $Q_\beta[4, 4] = Q_\beta[5, 5] = 0$, with an additional inequality constraint on $\bm{\xi}_\beta$ similar to the joint limit constraint.
The placing joint velocity $\qd_\beta$ only appears in the constraints.
The first and second constraints require the relative velocity between the hand and the object to be identical for pick and place in order to maintain that the relative hand-object transform for pick and place are the same at all timesteps.
The equation is asymmetric because ${_oT}_\alpha$ is constant.
Note that $R \inR{3\times3}, \bm{p} \inR{3}$ is the rotational and translational part of $T$, and $J^w_i, J^v_i \inR{3\times n}$ is the upper and lower half of $J_i$.

Even after successfully denoising $\bm{q}_i$, we frequently observed that the following trajectory generation fails.
This is because the discussion above does not include the approaching motion.
If $_hT_i$ is already close to the limit of reachable space or obstacles, there may be no feasible trajectory to reach it without colliding with the target object.
We therefore introduce the approaching pose $^a\bm{q}_i$ and entry pose $^e\bm{q}_i$, with their forward kinematics denoted as $^a_hT_i, {^e_hT_i}$ and decision variable ${^a\qd_i}, {^e\qd_i}$ respectively.
At each timestep, we add a cost to maintain the following conditions: (i) the approach hand pose seen from grasp pose ${_hT_i^{-1}}{^a_hT_i}$ has an identity rotation and a predefined translation (in this work, 10 cm in the hand approaching direction), (ii) the approach hand pose and the entry hand pose have the same rotation with a constant vertical offset (in this work, 20 cm).

While there are a large number of works in motion planning, they typically assume that all obstacles are known.
However, this is not the case when grasping unknown objects placed in a cluttered way based on partial observation.
We therefore take a more conservative approach where the hand linearly moves down from ${^e_hT}$ to ${^a_hT}$ and moves forward from ${^a_hT}$ to ${_hT}$.
After closing the hand, we lift it vertically to the same height as ${^e_hT}$, denoting it as ${^l_hT}$.
A dense joint trajectory is then acquired by linearly interpolating ${^e_hT}, {^a_hT}, {_hT}, {^l_hT}$ in the Cartesian space and solving Diff-IK.
We also apply this process for the placement, in reverse order.
As ${^l_hT}$ is relatively safe from unobserved obstacles, we use a PRM~\cite{prm}-based motion planner to connect between ${^l_hT_\alpha}$ and ${^l_hT_\beta}$. 

The overall denoising process is summarized as follows.
We start from multiple random $\bm{q}_i(K)$ based on heuristics.
We use $\bm{q}_i(K)$ as the initial value of $^a\bm{q}_i(K), {^e\bm{q}_i(K)}$ also.
At each timestep $k$, we solve the forward kinematics to get the hand pose ${_hT_\alpha}(k)$, which is fed into the network.
Based on the network output ${_h\bm{\xi}_\alpha(k)}$, we solve the optimization problem to get $\qd_i(k), {^a\qd_i(k)}, {^e\qd_i(k)}$ to iteratively update the joint configuration.
After the iterations from $k=K$ to $k=0$, we project ${_hT_\alpha}(0)$ to the grasp validity volume output by the network to filter out invalid grasps, and use the gravity rejection score volume to sort for the best one.
Finally, we compute the dense joint trajectory and send it to the robot.

In this work, we implement the entire algorithm in Python.
We use pydrake~\cite{drake} to parse the kinematics chain and detect collisions.
We solve QP via Drake's mathematical programming interface, which internally uses the Clarabel~\cite{clarabel} solver in this case.

\section{Experiment and Evaluation}
This section evaluates our approach using simulation and real hardware with the same problem setting.
A robotiq 2F-85 gripper~\cite{robotiq} is attached to a Franka-Emika Panda Robot~\cite{panda} located in front of a table.
In each scene, one to five unknown household objects are randomly arranged inside a 30 cm square region on the table surface.
In each trial, the robot picks one object and places it upright on a small rack, i.e., we allow arbitrary rotation around the z-axis.
We follow the metrics proposed in \cite{vgn} where the success ratio (SR) represents the number of successful grasps divided by the number of total trials, and the clear ratio (CR) represents the number of successful grasps divided by the total number of objects.
However, since we also target placement, we count a trial as success \textit{only if the object is placed upright with less than 5 cm position error.\footnote{
When the placement is successful, the position error is tyically less than 1 or 2 cm. The tolerance of 5 cm here is simply to reject cases where the object is dropped before placing.
}}
Similarly to the case of \cite{vgn}, each scene is terminated after two consecutive pick-place or detection failures.

\subsection{Comparison with Baseline}
To our best knowledge, there are no prior works that share the same problem setting with us.
We therefore create a baseline based on a naive depth-first search pipeline.
We select \cite{pvgn_bingham} as the grasp detector because our scene encoder and decoder are almost the same as the one in \cite{pvgn_bingham}, and it also considers power grasping by the Robotiq gripper.
This feedforward grasp detector detects all feasible grasp poses in the scene in the gravity rejection score order.
For each of the grasp poses, we first check whether it grasps the target object by checking the intersection between the hand region with the masked point cloud.
We then try to create the pick trajectory based on the interpolation of ${^e_hT_\alpha}, {^a_hT_\alpha}, {_hT_\alpha}, {^l_hT_\alpha}$, where ${_hT_\alpha}$ is the detection output.
Since we allow arbitrary yaw rotation for placing, we divide the yaw by 12.
For each of the placement candidates, we try to interpolate place trajectory by ${^l_hT_\beta}, {_hT_\beta}, {^a_hT_\beta}, {^e_hT_\beta}$, where ${_hT_\beta}$ is derived so that the relative transformation between the hand and the object is kept for pick and place.
If this process fails at any stage, we move to the next grasp pose until we find the first feasible trajectory for the robot to execute.

\begin{figure}[tbp]
    \centerline{\includegraphics[width=0.8\columnwidth]{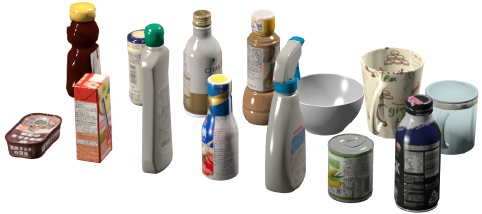}}
    \caption{
        Household object meshes for simulation evaluation.
        }
    \label{fig:sim_mesh}
\end{figure}

For simulation, we use Drake~\cite{drake} with the hydroelastic contact model~\cite{hydroelastic_2019} and the SAP solver~\cite{drake_sap}.
We use the ground truth object poses, and select the one with the highest centroid as the target.
The segmentation mask is also rendered based on the ground truth.
Once the target is placed on the rack, it is removed from the simulation.
Figure~\ref{fig:sim_mesh} shows the 14 object meshes we used for the evaluation.

\begin{figure}[tbp]
    \centerline{\includegraphics[width=0.9\columnwidth]{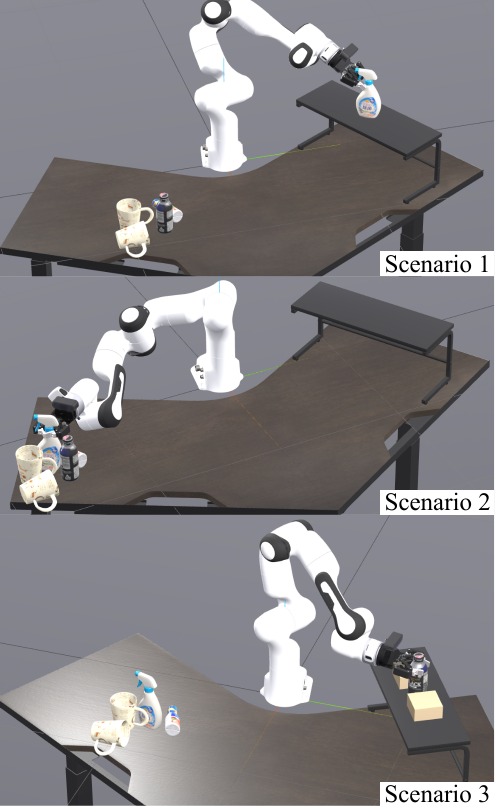}}
    \caption{
    Three scenarios for the robot to pick objects from the table top and place them on the small rack standing upright.
    (i) Both the pick and place location are in the middle of the robot's reachable space.
    (ii) Picking is hard because the objects are far from the robot.
    (iii) Placing is hard due to two obstacles next to the desired placement location.
    }
    \label{fig:scenarios}
\end{figure}

In order to investigate the effect of different difficulties for pick and place locations, we create three scenarios:
(i) both the pick and place locations are in the middle of the robot's reachable space; (ii) objects to pick are located far from the robot so that they are hard to reach; (iii) two obstacles are placed at both sides of the target place location so that placing is hard.
Figure~\ref{fig:scenarios} shows simulation captures of these three scenarios.

Table~\ref{tab:pvgn} summarizes the result.
In most cases, our approach outperformed the baseline in both SR and CR, presumably because our approach combines two paths, feedforward and diffusion, while the baseline only has one.
The larger contribution, though, lies on the computation time side, as shown in the third row.
Even though the baseline typically took only 10 ms for inference, the whole pipeline required a significantly longer time to search for a valid pick-place trajectory.
In the 4th and 5th row, we show the number of trials required to solve the pick and place trajectory.
When picking was difficult (scenario 2), finding the pick trajectory required more than three times more trials than in scenario 1.
When placing was difficult (scenario 3), finding the place trajectory required more than 10 times more trials.
In the 6th row of the table, we show the time spent on the trajectory generation only, proving that this part dominates the computation time of the whole pipeline.
In the case of our approach, the inference typically took more than 40 ms for the total of initial conditioning and 100 denoising iterations.
While this is more than four times longer than the baseline, the computation time as a whole pipeline is less than half in scenario 1.
Furthermore, the increase of computation time with the difficulty was less, resulting in a 2.7 times faster execution in scenario 2 and a 7.2 times faster execution in scenario 3.
This result indicates the high searching efficiency of our approach, which always guarantees that it stays in the reachable and collision-free manifold.

\begin{table}[htbp]
    \caption{Score and computation time of three scenarios}
    \begin{center}
    \begin{tabular}{lcccccc}
    \Xhline{3\arrayrulewidth}
    \\[-1em]
    &\multicolumn{2}{c}{Scenario 1} & \multicolumn{2}{c}{Scenario 2} & \multicolumn{2}{c}{Scenario 3}\\
    & Diff. & F.F. & Diff. & F.F. & Diff. & F.F. \\
    \Xhline{2\arrayrulewidth}
    Success Ratio [\%] & \bfseries{79.9} & 76.0 & \bfseries{71.2} & 67.9 & \bfseries{75.2} & 70.1 \\
    Clear Ratio [\%] & \bfseries{74.1} & 71.4 & 49.1 & \bfseries{49.7} & \bfseries{70.2} & 64.2 \\
    Time Elapsed [s] & \bfseries{1.6} & 3.6 & \bfseries{2.0} & 5.4 & \bfseries{1.9} & 13.7 \\
    \# Solve Pick Traj. & - & 60 & - & 202 & - & 102 \\
    \# Solve Place Traj. & - & 21 & - & 21 & - & 260 \\
    Traj. IK Time [s] & - & 3.3 & - & 4.8 & - & 13.3 \\
    \Xhline{3\arrayrulewidth}
    \end{tabular}
    \label{tab:pvgn}
    \end{center}
\end{table}

\begin{figure}[tbp]
    \centerline{\includegraphics[width=1.0\columnwidth]{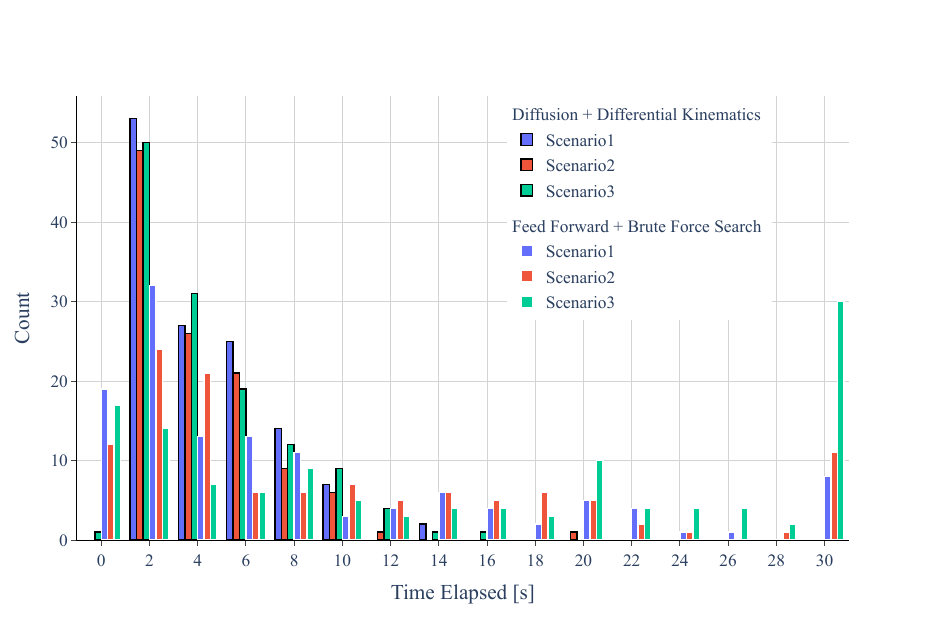}}
    \caption{
    Computation time histogram of the three scenarios.
    In either scenario, the baseline had a large diversity of computation time, while in our case, it was more consistent and predictable.
    }
    \label{fig:time_elapsed}
\end{figure}

In Fig.~\ref{fig:time_elapsed}, we plot a histogram of the computation time shown in Table~\ref{tab:pvgn}.
In either scenario, the baseline had a large diversity of computation time.
Even though some trials took less than one second, many exceeded more than 10 seconds.
This unpredictability is non-preferrable in real-world applications.
In our case, on the other hand, the computation time was more consistent and predictable.
Note that even in our approach, there is a fluctuation due to the random initial guess, however, its diversity is limited compared to the baseline.

As mentioned above, we implemented the algorithm in Python.
Switching to languages with a faster iteration, such as C++, can considerably boost the speed.
In addition, as we didn't introduce any parallelization except for the neural network inference, parallelizing the trajectory computation will also provide a notable impact.
At the same time, such improvements would benefit both the baseline and our approach equally, making the comparison in this section general among different implementations.

\subsection{Ablation Study}
This section provides an ablation study to investigate the contribution of each component.
Table~\ref{tab:ablation} summarizes the result in the score order.
The simulation setting is identical to scenario 1 in the previous section.
In case 2, we disabled selecting the best pick-place trajectory from the four candidates in the same batch based on the gravity rejection score volume, but randomly selected one.
This resulted in a decrease of 6.8\% SR and 2.7\% CR, showing that sorting generated grasp based on gravity rejection score is effective.
In case 3, we disabled removing candidates by the value of grasp validness volume.
This resulted in a larger decrease of 11.2\% SR and 3.9\% CR from the full version in case 1.
This result is reasonable: since our denoising process is constrained by the differential kinematics, the hand pose may not converge to a valid grasp pose, which is a more critical problem than selecting a valid yet low-score grasp.

As shown in case 4, conditioning with the gravity rejection score bin has the most significant effect on the score, with 11.9\% SR and 10.5\% CR.
Qualitatively, we observed that during the denoising process, the hand stops moving toward the object once it reaches ``close enough" to it, even though there are still remaining timesteps and moving even closer to the object will result in a more robust grasp.
Without conditioning by the gravity rejection score bin, it results in barely-able-to-grasp poses.
With a small detection error, the object may be ejected from the hand due to the too-shallow hand insertion; even if the grasp is successful, it is fragile against disturbance.
When conditioned with a high gravity rejection score bin, the hand stops only when it's close enough for \textit{robust} grasps, which are typically power grasps or grasping close to the center of mass.

Case 5 and case 6 had identical inference pipelines: neither the grasp validness volume nor the gravity rejection score volume was used.
However, it differs in the training phase, where in case 5 the two volumes were jointly trained with the diffusion model while in case 6 they were not.
The score difference of 3.9\% SR and 4.3\% CR indicates that even if not used at the inference time, jointly training a feedforward grasp detection network as a secondary task is beneficial for training a grasp diffusion network, presumably because it guides a better scene representation.

\begin{table}[htbp]
    \caption{Simulation result of ablation study}
    \begin{center}
    \begin{tabular}{l@{\hskip 6pt}cccccc}
    \Xhline{3\arrayrulewidth}
    \\[-1em]
     & 1 & 2 & 3 & 4 & 5 & 6 \\
    \Xhline{2\arrayrulewidth}
    Cond. by Grav. Rej. Sco. & \cmark & \cmark & \cmark & \xmark & \cmark & \cmark \\
    Train F.F. as 2nd Task & \cmark & \cmark & \cmark & \cmark & \cmark & \xmark \\
    Grasp Valid. Mask & \cmark & \cmark & \xmark & \cmark & \xmark & \xmark \\
    Sort by Grav. Rej. Sco. & \cmark & \xmark & \cmark & \cmark & \xmark & \xmark \\
    \Xhline{2\arrayrulewidth}
    Success Ratio [\%] & 79.9 & 73.1 & 68.7 &  68.7 & 66.1 & 62.2 \\
    Clear Ratio [\%] & 74.1 & 71.4 & 70.2 & 63.6 & 65.1 & 60.8 \\
    \Xhline{3\arrayrulewidth}
    \end{tabular}
    \label{tab:ablation}
    \end{center}
\end{table}

\subsection{Real Robot Verification}
\begin{figure}[tbp]
    \centerline{\includegraphics[width=0.8\columnwidth]{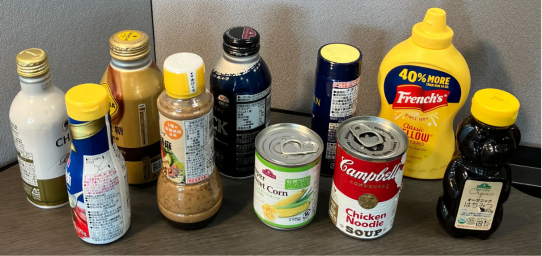}}
    \caption{
        Objects used for real robot validation.
        All items except the mustard bottle were full of ingredients.
    }
    \label{fig:real_objects}
\end{figure}

This section provides a real-world verification of the proposed method.
We used a pair of external industry cameras and a learning-based stereo matching~\cite{learned_stereo} to acquire an RGB-D image.
We detected the objects by GroundingDINO~\cite{grounding_dino} and segmented them by SAM~\cite{sam}.
The 6D object poses are estimated by DiffusionNOCS~\cite{diffusion_nocs}.
Figure~\ref{fig:real_objects} shows the objects we used, and Fig.~\ref{fig:experiment} shows an example of the test.

Table~\ref{tab:real} summarizes the result, which is consistent with the simulation experiment where our approach outperformed the baseline in both SR/CR and computation time.
The computation times are longer because they include the object detection and pose estimation pipeline, while the 6-second reduction from the baseline to our approach is due to our contribution.
During the experiment, we observed that pose estimation error was one of the major sources of placement failure.
We therefore introduced a variant of our approach: we used another pair of cameras located near the placement location to re-identify the object's pose before the robot placed it.
We used the updated $_oT_\beta$ to solve the place trajectory again, which turned out to effectively compensate for the pose estimation error and the error due to the object's movement inside the hand.
We show its result in the rightmost column of the table.
Notice that the time for the refinement is not included in the table.

\begin{table}[htbp]
    \caption{Result of real robot validation}
    \begin{center}
    \begin{tabular}{cccc}
    \Xhline{3\arrayrulewidth}
    & Baseline & Ours & \makecell{Ours w/\\pre-place refinement}\\
    \Xhline{2\arrayrulewidth}
    \\[-1em]
    SR & 38/56 = 67.9\% & 43/61 = 70.5\% & 46/61 = 75.4\% \\
    CR & 38/60 = 63.3\% & 43/60 = 71.7\% & 46/60 = 76.7\%\\
    Time [s] & 17.7 & 11.7 & 10.9 \\
    \Xhline{3\arrayrulewidth}
    \end{tabular}
    \label{tab:real}
    \end{center}
\end{table}

\section{Limitation and Future Work}
\subsubsection{Failure Modes}
During the evaluations, we observed two major failure modes.
The first one is grasp failure due to inaccurate approach direction.
Since both the grasp validness volume and the gravity rejection score volume are based on the voxel location, they can only verify the grasp's translation but not its rotation.
Introducing an additional post-processor to evaluate SE(3) grasp poses will improve performance while it remains in our future works.
The other failure case is placement failure: if the object moves inside the hand during the pick-to-place trajectory, there is a high chance that it will fall over during the placement.
While the gravity rejection score represents the magnitude of gravity direction disturbance a grasp can resist, the gravity direction relative to the grasp varies during the pick-to-place trajectory execution.
Extending it to consider multiple disturbance directions is a possible extension.

\subsubsection{Local Minima at Denoising Process}
As another limitation, we occasionally observed that the hand was trapped in the concave-shaped part of the surrounding obstacles during denoising, especially when the scene is heavily cluttered.
This is because our collision avoidance is based on hard inequality constraints.
Relaxing the constraints by introducing potential fields or first reaching the hand without considering the collision and then repelling the colliding part from the obstacles may imporve the detection success ratio, while it still remains as future work.

\subsubsection{Computation Efficiency}
In this work, we implemented the kinematics loop in Python for the ease of combining with the diffusion model.
In order to align with this, we also implemented the baseline naively in Python, which is less performant than the state-of-the-art libraries such as MoveIt!~\cite{moveit}.
While we believe this is an orthogonal to our contribution, i.e., we can also introduce such optimization technics to our approach, actual validation still remains as our future work.


\section{Conclusion}
In this work, we proposed a framework for detecting dense pick-and-place joint trajectories for unknown objects in a cluttered scene.
The key idea was to train a diffusion model that estimates the noise in the form of spatial velocity, and feed it to a multi-task Diff-IK, so that the state is guaranteed to always stay in reachable and collision-free manifold.
We experimentally confirmed that our approach is highly computationally efficient compared with a depth-first-search-based baseline, especially when either pick or place is hard in reachability or collision-avoidance perspective.
We also showed that jointly training a diffusion-based grasp detector with a feedforward one and using the feedforward one's output to filter/sort the denoised grasps is highly effective in boosting performance.

\bibliographystyle{IEEEtran}
\bibliography{citations.bib}

\vspace{12pt}

\end{document}